\documentclass[accepted]{article}

\usepackage{tabularx}
\newcolumntype{C}{>{\Centering\arraybackslash}X} 
\usepackage{microtype}
\usepackage{graphicx}
\usepackage{xspace}
\usepackage{subfig}
\usepackage{booktabs} 
\usepackage[noend]{algpseudocode}
\usepackage{hyperref}
\usepackage{xurl}

\usepackage[most]{tcolorbox}
\usepackage[framemethod=tikz]{mdframed}

\definecolor{mybrown}{RGB}{128,64,0}  

\gdef\Sepline{
  \par\noindent\makebox[\linewidth][l]{%
  \hspace*{-\mdflength{innerleftmargin}}%
   \tikz\draw[thick,dashed,gray!60] (0,0) --%
        (\textwidth+\the\mdflength{innerleftmargin}+\the\mdflength{innerrightmargin},0);
  }\par\nobreak}

\usepackage{icml2024}

\usepackage{amsmath}
\usepackage{amssymb}
\usepackage{mathtools}
\usepackage{amsthm}
\usepackage{hyperref}       
\usepackage{url}            
\usepackage[capitalize,noabbrev]{cleveref}

\theoremstyle{plain}

\theoremstyle{definition}

\theoremstyle{remark}

\usepackage{enumitem}
\usepackage{tikz}
\usetikzlibrary{fit}

\pgfdeclarelayer{background}
\pgfsetlayers{background,main}
\usetikzlibrary{shapes,arrows}
\usepackage[textsize=tiny]{todonotes}

\usetikzlibrary{positioning}

\definecolor{myyellow}{RGB}{255,255,150}

\newcommand\mytext[3][\scriptsize]{#2\\#1 #3}
\newcommand\mynode[4][]{%
  \node[mynode,#1,text width=\the\dimexpr#2cm*3] (#3) {\mytext{#3}{#2 #4}}; 
}
\newcommand\mywidth{1cm}

\newcommand{\data}{TriMaster100\xspace}
\newcommand{\alg}{SSC-CoT\xspace}
\newcommand{\lm}{{\ttfamily{LLEMMA}}\xspace}
\newcommand{\sccot}{CoT-SC\xspace}

\usepackage{tikz}

\newcommand*\circled[1]{\tikz[baseline=(char.base)]{
            \node[shape=circle,draw,inner sep=1.4pt] (char) {#1};}}

\icmltitlerunning{Submission and Formatting Instructions for ICML 2024}

\begin{document}

\twocolumn[
\icmltitle{Stepwise Self-Consistent Mathematical Reasoning with Large Language Models}



\icmlsetsymbol{equal}{*}

\begin{icmlauthorlist}
\icmlauthor{Zilong Zhao}{yyy}
\icmlauthor{Yao Rong}{yyy}
\icmlauthor{Dongyang Guo}{yyy}
\icmlauthor{Emek Gözlüklü}{yyy}
\icmlauthor{Emir Gülboy}{yyy}
\icmlauthor{Enkelejda Kasneci}{yyy}
\end{icmlauthorlist}

\icmlaffiliation{yyy}{Technical University of Munich, Munich, Germany}

\icmlcorrespondingauthor{Zilong Zhao}{zilong.zhao@tum.de}
\icmlcorrespondingauthor{Yao Rong}{yao.rong@tum.de}

\icmlkeywords{Machine Learning, ICML}

\vskip 0.3in
]



\printAffiliationsAndNotice{}


\begin{abstract}
Using Large Language Models for complex mathematical reasoning is difficult,  primarily due to the complexity of multi-step reasoning. 
The main challenges of this process include (1)  selecting critical intermediate results to advance the procedure, and (2) limited exploration of potential solutions.
To address these issues, we introduce a novel algorithm, namely Stepwise Self-Consistent Chain-of-Thought (\alg). \alg employs a strategy of selecting intermediate steps based on the intersection of various reasoning chains. Additionally, \alg enables the model to discover critical intermediate steps by querying a knowledge graph comprising relevant domain knowledge. 
To validate \alg, we present a new dataset, \data, tailored for complex trigonometry problems. This dataset contains 100 questions, with each solution broken down into scored intermediate steps,  
facilitating a comprehensive evaluation of the mathematical reasoning process. 
On \data, \alg triples the effectiveness of the state-of-the-art methods. Furthermore, we benchmark \alg on the widely recognized complex mathematical question dataset, MATH level 5, and it surpasses the second-best method by 7.2\% in accuracy.
Code and the \data dataset can be found at: 
\url{https://github.com/zhao-zilong/ssc-cot}.
\end{abstract}

\vspace{-1em}
\section{Introduction}
\label{sec:intro}

Large Language Models (LLMs) are increasingly being utilized for mathematical reasoning tasks~\cite{imani2023mathprompter, NEURIPS2022_9d560961,kojima2205large,wang2022self, azerbayev2023llemma,chen2022program,xin2023lego,trinh2024solving,paranjape2023art}. However, these models predominantly focus on simpler mathematical problems~\cite{GSM8K,AQuA,MultiArith,SVAMP}.
Tackling complex mathematical questions remains a significant challenge~\cite{trinh2024solving,azerbayev2023llemma,xin2023lego}. 
This difficulty often arises from the foundational models' limited knowledge, impairing their ability to comprehend complex questions. Additionally, the long reasoning chains required for complex problem-solving necessitate carefully designed multi-step reasoning~\cite{yao2023tree,besta2023graph} methods to arrive at the final answer.

\begin{figure}[t]
\centering
\resizebox{\linewidth}{!}{
    \begin{tikzpicture}[font=\Large,thick, every text node part/.style={align=center}]
 
 
 
\node[draw, execute at begin node=\setlength{\baselineskip}{3ex},
    minimum width=0.5cm, fill=gray!5,
    minimum height=0.5cm, rounded corners=.2cm,
    inner sep=5pt] at (0, 0) (q1) {Q: Simplify $\tan{100^{\circ}} + 4\sin{100^{\circ}}$.};

\node[draw, execute at begin node=\setlength{\baselineskip}{3ex},
    minimum width=0.5cm, fill=gray!5,
    minimum height=0.5cm, rounded corners=.2cm,
    inner sep=5pt] at (8, 0) (q2) {Q: Simplify $\tan{100^{\circ}} + 4\sin{100^{\circ}}$.};
    
\node[draw,
    below = 2cm of q1, execute at begin node=\setlength{\baselineskip}{3ex},
    minimum width= 0.5cm,
    minimum height=0.5cm, rounded corners=.2cm,
    inner sep=5pt] at (0, -.5) (left-s1) {Try different identities.};

\node[draw,
    below = 4.5cm of left-s1, execute at begin node=\setlength{\baselineskip}{3ex},
    minimum width= 0.5cm,
    minimum height=0.5cm, rounded corners=.2cm,
    inner sep=5pt] at (0, -.5) (left-ans) {Answer: $-\sqrt{3}$};

\node (label) at (0,-7.6){};

\node[draw, fill=orange!10,rounded corners=3pt] (expl) at (0,-7.2)
  {Hit after 1000 attempts. \\Hard to choose helpful identities.};

\begin{pgfonlayer}{background}
    \node[fill=purple!5, fit=(q1)(label),rounded corners=0.1cm,inner sep=.2cm]{};
\end{pgfonlayer}

\node[draw,
    below = 0.5cm of q2, execute at begin node=\setlength{\baselineskip}{3ex},
    minimum width= 0.5cm,
    minimum height=0.5cm, rounded corners=.2cm,
    inner sep=5pt] at (8, -.5) (right-s1) {Try and \textbf{verify} different identities.};

\node[draw,
    below = 2.5cm of right-s1, execute at begin node=\setlength{\baselineskip}{3ex},
    minimum width= 0.5cm,
    minimum height=0.5cm, rounded corners=.2cm,
    inner sep=5pt] at (8, -.5) (right-s2) {Get \textbf{Critical} identities.};

\node[circle, draw, fill=blue!10,
    below = 1.5cm of right-s1, execute at begin node=\setlength{\baselineskip}{3ex},
    minimum width= 0.1cm,
    minimum height=0.1cm, 
    inner sep=1pt] at (11, -.5) (kg) {KG};


\node[draw,
    below = 4.5cm of right-s1, execute at begin node=\setlength{\baselineskip}{3ex},
    minimum width= 0.5cm,
    minimum height=0.5cm, rounded corners=.2cm,
    inner sep=5pt] at (8, -.5) (right-ans) {Answer: $-\sqrt{3}$};

\node (label2) at (8,-7.6){};

\node[draw, fill=green!10,rounded corners=3pt] (expl) at (8,-7.2)
  {Hit after 50 attempts. \\ Critical identities are available.};

\begin{pgfonlayer}{background}
    \node[fill=lime!5, fit=(q2)(label2),rounded corners=0.1cm,inner sep=.2cm]{};
\end{pgfonlayer}

\draw[-latex]  (q1) edge (left-s1);
\draw[-latex]  (left-s1) edge (left-ans);

\draw[-latex]  (q2) edge (right-s1);
\draw[-latex]  (right-s1) edge (right-s2);
\draw[-latex] (right-s2) edge (right-ans);
\draw[-latex]  (kg) edge[bend left] (right-s2.east);

 
\end{tikzpicture}}
    		\caption{Our \alg (Right) improves the ability of LLMs (Left) to solve complex mathematical questions.} 
		\label{fig:teaser}
  \vspace{-1em}
\end{figure}
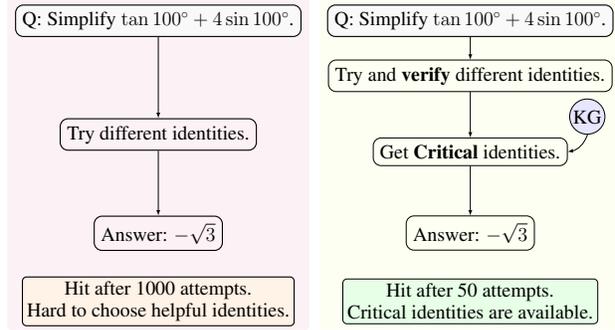

Although existing multi-step reasoning algorithms demonstrate notable capabilities in solving mathematical problems, they encounter challenges when addressing more complex mathematical questions. Identifying critical intermediate steps is important for guiding the model towards correct solutions in these difficult problems. However, existing approaches are not effective in discovering these critical intermediate steps. Additionally, the model tends to become stuck at an intermediate step, hindering further progress. To address these two challenges, we propose a novel framework named Stepwise Self-Consistent Chain-of-Thought (\alg), which enhances LLMs' capabilities in multi-step reasoning for complex mathematical questions. 

The design of \alg is inspired by the scenario where humans tackle a complex problem. They often go through multiple attempts, reaching a promising intermediate step where they may encounter a roadblock. At this moment, a hint can facilitate their continued progress. In practice,
\alg first selects a set of potential intermediate steps by determining their intersection across multiple reasoning chains. It then evaluates the correctness of each step within this intersection set and selects the most optimal ones for progression. To overcome being stuck at a step, it also queries a Knowledge Graph (KG) that contains external mathematical knowledge. The information gathered, along with identified key intermediate steps, is then crafted into hints that guide and further prompt the model. Our algorithm enhances the model's capability by providing them with critical intermediate steps, thereby enabling it to discover important steps it would not have identified independently. As illustrated in \Cref{fig:teaser}, we present a trigonometry question to ChatGPT 4 (on the left), and it arrives at the correct answers after a thousand attempts\footnote{Example question can be found at \url{https://openai.com/research/improving-mathematical-reasoning-with-process-supervision}.}. \alg (on the right) helps the model identify critical intermediate steps through verification and KG integration, enabling the model to reach the correct answer within 50 attempts.

To fairly validate the effectiveness of multi-step reasoning methods in discovering critical intermediate steps, we collected a dataset, \data, comprising 100 complex trigonometry questions (up to Mathematical Olympiad difficulty level). Each question is accompanied by a solution divided into intermediate steps, each of which is scored. Recognizing the challenge for LLMs to solve complex mathematical questions within a limited number of attempts, \data focuses on evaluating the model's ability through the scoring of intermediate steps. This differs from existing mathematical question benchmarks, which typically concentrate only on the correctness of the final answer, without considering the evaluation of these intermediate phases. 
{We posit that evaluating intermediate results is crucial for complex mathematical questions. Given the typically low accuracy of all mathematical reasoning algorithms on such datasets, distinguishing the true capabilities of these algorithms based solely on final outcomes is inappropriate.}
Beyond this dataset, we benchmark our model on MATH level 5~\cite{math}, which is recognized as a complex dataset, to showcase its ability in solving various types of mathematical questions.   
To summarize, our contributions are as follows: 
\vspace{-8pt}
\begin{itemize}[itemsep=0.05mm, leftmargin=*]
    \item We introduce a novel multi-step reasoning algorithm \alg to tackle complex mathematical questions. This algorithm significantly improves LLM's capabilities in identifying critical intermediate steps for problem-solving.
    \item We propose a procedure to establish a KG and allow LLM to efficiently retrieve information in textual form, facilitating the generation of critical intermediate results.
    \item We provide a new dataset named \data designed for evaluating intermediate results in very complex mathematical questions. 
    \item We benchmark \alg with state-of-the-art (SOTA) multi-step reasoning algorithms on \data and MATH datasets, where our algorithm significantly surpasses others, demonstrating its ability in solving complex questions by identifying critical intermediate steps.
\end{itemize}



\vspace{-1em}
\section{Related Work}
\label{sec:related}

\paragraph{LLMs for Mathematical Reasoning.}

To enhance mathematical reasoning capabilities, recent efforts can be categorized into two approaches: one is training domain-specific models such as {\ttfamily{LLEMMA}}~\cite{azerbayev2023llemma}, while the other leverages in-context learning methods without training. The former method enhances model capabilities but demands substantial resources. 
Furthermore, this specialization may lead to a reduced ability to understand broader contexts, potentially compromising the models' effectiveness in comprehending and solving questions outside the fine-tuning dataset. 
Therefore, our focus is on an innovative in-context learning approach. A representative of such an approach is Chain-of-Thought (CoT)~\cite{kojima2205large,NEURIPS2022_9d560961}. CoT significantly improves the problem-solving abilities of Large Language Models (LLMs) in mathematics by incorporating intermediate steps into their outputs.
Following this principle, Tree-of-Thought (ToT)~\cite{yao2023tree,long2023large} and Graph-of-Thought (GoT)~\cite{besta2023graph} prompt the model to evaluate the current step and choose the next promising step. 
However, ToT and GoT have a limitation: they consider only one step ahead without emphasizing the overview of the problem, which can slow down the problem-solving process.
Our method addresses this by generating the next step based on the overlapping intermediate steps from multiple trials (chains).
CoT with Self-Consistency (\sccot) proposed by \citet{wang2022self} similarly leverages the overlap between multiple chains, Nevertheless, it restricts this overlap selection to the final answers. In contrast, our approach focuses on the correctness of intermediate results, leading to improved reasoning performance compared to \sccot.

\vspace{-0.5em}
\paragraph{Retrieval Augmentation for Mathematical Reasoning}
The Retrieval-Augmented Generation (RAG)~\cite{NEURIPS2020_6b493230} technique was first proposed for knowledge-intensive NLP tasks. The framework contains a retrieval component which can fetch relevant information from a structured knowledge base, such as database or knowledge graph. This framework has been successfully implemented for commonsense reasoning~\cite{ragc}, and middle-school algebra and geometry QA~\cite{levonian2023retrieval}. 
{Studies \cite{xin2023lego,yang2023leandojo} have explored the integration of RAG with LLMs for mathematical reasoning, specifically for theorem proving. These works utilize a library containing vector embeddings of definitions and theorems for retrieval using cosine similarity based on the question's embedding.
In this paper, our approach, utilizing a knowledge graph, diverges by focusing on providing relevant trigonometry identities directly linked to the question's elements, such as trigonometric functions and angles, enabling a more direct match than the cosine similarity method used in \cite{xin2023lego,yang2023leandojo} for theorem retrieval. This ensures a clearer relation between the question and the information retrieved.}

\vspace{-0.5em}
\section{\data Dataset}
\label{sec:dataset}
{Existing datasets for complex mathematical questions include only the final answer or encapsulate the entire reasoning process in a single string, lacking clear intermediate results.  
These datasets fall short in adequately distinguishing the nuanced problem-solving abilities required for such tasks, 
because of the low accuracy current mathematical reasoning algorithms achieve on complex questions. 
Relying solely on accuracy for evaluation proves inadequate in these challenging contexts. To address this, we introduce the \data dataset, comprising 100 challenging trigonometry questions ranging from senior high school to Mathematical Olympiad levels. \data's innovative evaluation methodology not only assesses final answer accuracy but also scores intermediate results, enabling a more detailed and accurate evaluation of an algorithm's problem-solving process. This fills a critical gap in assessing complex mathematical reasoning.
}

\begin{figure}
\centering
\resizebox{0.98\linewidth}{!}{
    \begin{tikzpicture}[font=\Large,thick, every text node part/.style={align=center}]
 
 
 
\node[draw, execute at begin node=\setlength{\baselineskip}{3ex},
    minimum width=5cm,
    minimum height=2cm,
    inner sep=6pt] at (0, 0) (q) {\textbf{Question:} \\ If $\alpha \in (0,\frac{\pi}{2})$, $\beta \in (0,\frac{\pi}{2})$, and $\tan{\alpha}=(1+\sin{\beta})/\cos{\beta}$, \\find the value of $2\alpha-\beta$. };

\node[draw,
    below left=of q, execute at begin node=\setlength{\baselineskip}{3ex},
    minimum width= 4cm,
    minimum height=2cm,
    inner sep=6pt] at (-1, -.5) (s1) {\textbf{Step 1 (1 point):}\\ $1+\sin{\beta}=1+2\sin{\frac{\beta}{2}}\cos{\frac{\beta}{2}}$};

\node[draw, execute at begin node=\setlength{\baselineskip}{3ex},
    below = .5cm of s1,
    minimum width= 4cm,
    minimum height=2cm,
    inner sep=6pt] (s2) {\textbf{Step 2 (2 points):}\\ $1+2\sin{\frac{\beta}{2}}\cos{\frac{\beta}{2}} = (\sin{\frac{\beta}{2}} + \cos{\frac{\beta}{2}})^2$};

\node[draw, execute at begin node=\setlength{\baselineskip}{3ex},
    below right=of q,
    minimum width= 4cm,
    minimum height=2cm,
    inner sep=6pt] at (1, -.5) (s3) {\textbf{Step 3 (1 point):}\\ $1+\sin{\beta}=1+2\sin{\frac{\beta}{2}}\cos{\frac{\beta}{2}}$};

\node[draw, execute at begin node=\setlength{\baselineskip}{3ex},
    below = .5cm of s3,
    minimum width= 4cm,
    minimum height=2cm,
    inner sep=6pt] (s4) {\textbf{Step 4 (2 points):}\\ $\cos^2{\frac{\beta}{2}} - \sin^2{\frac{\beta}{2}} = (\cos{\frac{\beta}{2}} + \sin{\frac{\beta}{2}}) (\cos{\frac{\beta}{2}} - \sin{\frac{\beta}{2}})$};

\node[draw, execute at begin node=\setlength{\baselineskip}{3ex},
    below = 5.5cm of q,
    minimum width= 4cm,
    minimum height=2cm,
    inner sep=6pt] (s5) {\textbf{Step 5 (5 points):}\\ $(1+\sin{\frac{\beta}{2}})/\cos{\beta} = (\sin{\frac{\beta}{2}} + \cos{\frac{\beta}{2}}) /  (\cos{\frac{\beta}{2}}) - \sin{\frac{\beta}{2}})$};

\node[draw, execute at begin node=\setlength{\baselineskip}{3ex},
    below = .5cm of s5,
    minimum width= 4cm,
    minimum height=2cm,
    inner sep=6pt] (s6) {\textbf{Step 6 (6 points):}\\ $ (\sin{\frac{\beta}{2}} + \cos{\frac{\beta}{2}}) /  (\cos{\frac{\beta}{2}}) - \sin{\frac{\beta}{2}}) = (\tan{\frac{\beta}{2}}+1) / (1-\tan{\frac{\beta}{2}})$};
 
\node[draw, execute at begin node=\setlength{\baselineskip}{3ex},
    below = .5cm of s6,
    minimum width= 4cm,
    minimum height=2cm,
    inner sep=6pt] (s7) {\textbf{Step 7 (7 points):}\\ $  (\tan{\frac{\beta}{2}}+1) / (1-\tan{\frac{\beta}{2}}) = \tan{(\frac{\beta}{2} + \frac{\pi}{4})}$};
 
\node[draw, execute at begin node=\setlength{\baselineskip}{3ex},
    below = .5cm of s7,
    minimum width= 4cm,
    minimum height=2cm,
    inner sep=6pt] (s8) {\textbf{Step 8 (8 points):}\\ $ (\frac{\beta}{2} + \frac{\pi}{4}) = \alpha$,\\ $2\alpha - \beta = \frac{\pi}{2}$};
 
 

\draw[-latex]  (q) edge (s1);
\draw[-latex]  (s1) edge (s2);
\draw[-latex]  (s2) edge (s5);

\draw[-latex]  (q) edge (s3);
\draw[-latex]  (s3) edge (s4);
\draw[-latex]  (s4) edge (s5);

 \draw[-latex]  (s5) edge (s6);
\draw[-latex]  (s6) edge (s7);
\draw[-latex]  (s7) edge (s8);
 
\end{tikzpicture}}
    		\caption{Example of the annotated intermediate steps with its scores in \data.} 
       \vspace{-1.1em}
		\label{fig:question}
\end{figure}
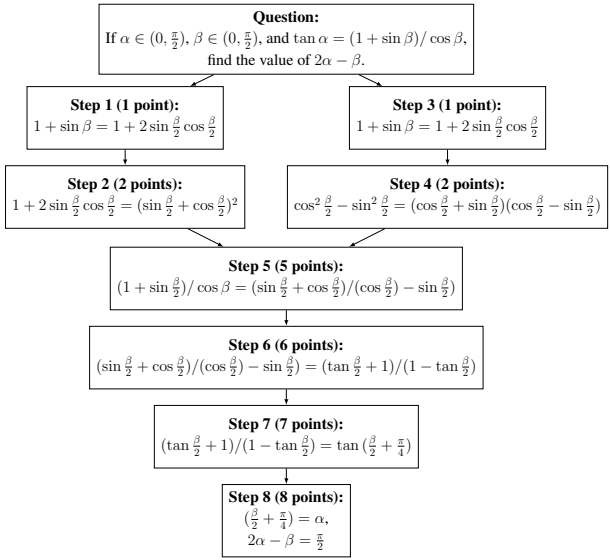

\begin{figure*}[t]
	\begin{center}
			\includegraphics[width=0.95\textwidth]{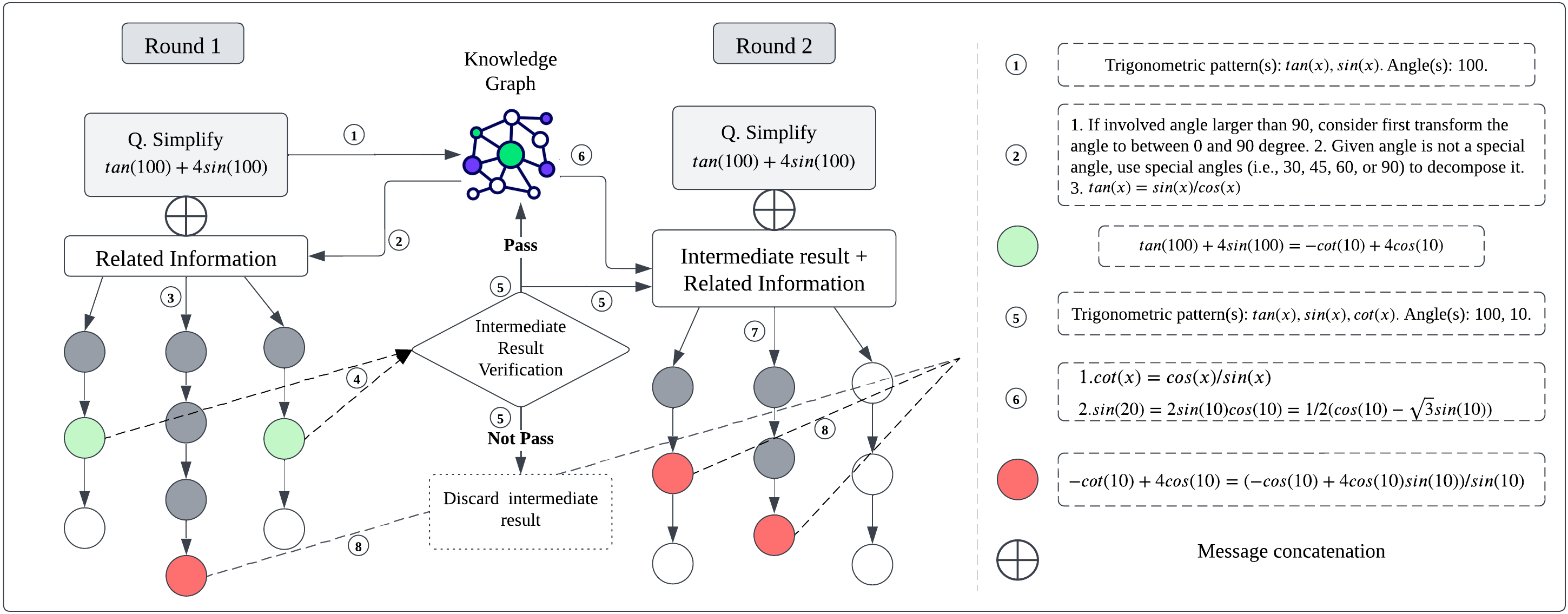}
   \vspace{-0.5em}
		\caption{An example of Stepwise Self-Consistent Chain-of-Thought workflow.} 
		\label{fig:workflow}
 	\end{center}
  \vspace{-1em}
\end{figure*}
\subsection{Dataset Construction}
To ensure the complexity of the questions, we employed GPT-4 as a standard, presenting each question to it three times as a prompt. A question was defined as \textit{complex} if GPT-4 failed to solve it completely on at least one occasion. 
Notably, if GPT-4 delivered a direct answer without explaining the reasoning process, we explicitly requested the model to outline the intermediate steps involved (prompts in Appendix~\ref{app:gpt}). A question is considered solved only if both the intermediate steps and the final answer provided are correct. {Also due to the above reason, the Wolfram plugin in GPT-4~\cite{WMathematica}, which is a powerful computation plugin, is not used since it cannot output all reasoning steps.}
The resulting dataset comprises questions broken down into various intermediate steps, with each step being assigned a specific score. The aggregation of scores for all these steps forms an extensive evaluation, culminating in a total of 750 points across the 100 questions.
Comparing to prior math datasets, \data offers a thorough assessment of a model's reasoning capabilities by evaluating not just the final answer but also the intermediate steps. This approach is grounded in two fundamental reasons:
(1) Current models often fail to fully solve the problem, thereby not reaching the conclusive answer; (2) A correct final answer, if derived through incorrect intermediate steps, should not be considered as a demonstration of valid mathematical reasoning.

Figure~\ref{fig:question} illustrates an example of the annotated intermediate steps with their scores in the \data dataset. This question encompasses 8 intermediate results, resulting in total 8 points if the question is successfully solved. We recognize the possibility of parallel solution paths, e.g., step 1,2 and step 3,4. In such cases, scores for each distinct path should be calculated independently, and the final scores from these different paths are then summed up. It is important to understand that mathematical questions often allow multiple valid solutions. Therefore, even if a reasoning process correctly reaches step 6 without adhering to steps from 1 to 5, we still attribute a final score of 6 if all its previous steps are verified to be correct. Please note that \data includes labeled final results, which allows users to employ \data as prior math datasets to evaluate the accuracy of final answer. 

Trigonometry was selected due to its perceived difficulty and abstract nature~\cite{sayster2023high}, making it ideal for complex mathematical question reasoning. 
Looking ahead, we plan to integrate a KG to enhance mathematical reasoning abilities. Trigonometry's relatively limited fundamental identities facilitate the creation of a concise yet detailed KG.
This strategic focus on trigonometry ensures agility in developing, validating, and refining methodologies. The \data was independently audited by three researchers to ensure its accuracy.
\subsection{Human-Level Performance}

We offer a preliminary yet informative comparison to human-level performance by randomly selecting $10$ questions from \data and testing them with human participants. We recruited a group of five participants, each holding a Master's Degree. During the evaluation, each participant was given a maximum of one hour to solve two questions. Collectively, the five participants achieved 17 points out of a possible 63 for the ten questions. On average, participants spent 9 minutes per question, indicating that they did not develop new thoughts on solving the question after that time. Only one question was completely solved. The best-performing participant scored 9 points out of 13, while two participants scored 0 for their questions. This result indicates that questions in \data{} are difficult.

\section{Stepwise Self-Consistent Chain of Thought}
\label{sec:method}

In this section, we introduce the workflow of \alg, followed by the detail of the two core components in our algorithm: (1) The design of the KG in the context of trigonometry questions and the approach to retrieve information from it (\Cref{ssec:kg}); (2) The procedure of selecting critical intermediate steps (\Cref{ssec:overlap}).

\subsection{\alg Workflow}
Figure~\ref{fig:workflow} presents the \alg workflow for two rounds of trigonometry question querying. 
First, when presented with a question $Q$, step \circled{1} extracts a set of key information from the question, which is represented as $\mathcal{V} = E(p_{\theta},Q)$ where $E(\cdot)$ represents the extraction function and $p_{\theta}$ denotes an LLM with parameter $\theta$. With a set of extracted information $\mathcal{V}$, \alg queries related information from the KG for the round $k$, represented as $r_k = S(\mathcal{G},\mathcal{V})$ where $S(\cdot)$ represents the searching function and $\mathcal{G}$ denotes a KG (detail in Section~\ref{ssec:kg}). In step \circled{2}, the information $r_1$ ($k=1$), used as a \textit{hint} with the question $Q$, forms a prompt. This then leads to \circled{3}, which generates $N$ reasoning chains: $\mathcal{C}_i = G_1(p_{\theta},Q,r_1)$ where $i \in [1,N]$, and $G_1(\cdot)$ is the function to generate reasoning chain for round 1 with $N=3$ in this example, $G_1(\cdot)$ prompt template is added in Appendix~\ref{app:4}. 
Each circle in the chain symbolizes an intermediate result, which is defined as a \textbf{state} $x_i^j$, with $i$ and $j$ referring to $i$th chain position $j$. 

In the first round, all generated states are used to identify those states that share identical mathematical meanings (indicated in green). Results preceding these overlapping states within the same reasoning chain are marked as \textbf{Inactive} (shown in grey), signifying their exclusion in subsequent overlap searches.
Intermediate result selection detail is provided in Section~\ref{ssec:overlap}.
The selected overlapping states form a set and denotes as $\mathbf{S}$. 
Step \circled{4} involves sending $\mathbf{S}$ along with the original question $Q$ to a verifier. This is represented as $\mathbf{S}_v = \left\{ x_i^j \in \mathbf{S} \mid V(Q, x_i^1x_i^2...x_i^j) = 1 \right\}$, where $V$ denotes the verification function. 
It is important to note that, due to the identical nature of overlapping intermediate results, verification only requires a single chain of the result up to the intermediate result, rather than from all chains. In the current implementation of \alg, this verification role $V(\cdot)$ is fulfilled by the language model, its prompt template is added in Appendix~\ref{app:v}. In step \circled{5}, the intermediate results that do not pass the verification are discarded, while the ones that pass the verification are combined to a verified set of states $\mathbf{S}_v$. This set is used to query related information from the KG, i.e., {$r_k = S(\mathcal{G},E(p_{\theta},\mathbf{S}_v))$}. 
This information $r_k$, combined with the question and intermediate result, serves as a new prompt for the subsequent round of querying.
Step \circled{7} generates reasoning chains for round 2, i.e., $\mathcal{C}_i = G(p_{\theta},Q,r_2,\mathbf{S}_v)$ with $k=2$, prompt template of $G$ is added in Appendix~\ref{app:ssc_g}. In this round, we evaluate not only the results from the current round but also those from round 1 that were not marked Inactive. The selection of a new overlapping result (orange circle) in step \circled{8} triggers a repetition of steps \circled{4} through \circled{7}.
The algorithm proceeds until it reaches a predefined number of rounds or queries. To conclude the final result from \alg, we will first find out which intermediate result contains the conclusion statement. The final result is then derived from the majority vote among all these conclusion statements.

\begin{figure}[t]
	\begin{center}
        \resizebox{.8\linewidth}{!}{
        \begin{tikzpicture}[
node distance=2cm, every text node part/.style={align=center}, font=\tiny
]
\node[circle, draw, inner sep=1pt, fill=purple!10](1){$\sin^3{x}$};
\node[circle, draw, above = .5\mywidth of 1, inner sep=1pt, fill=cyan!20](2){$\sin{3x}=$  \\$ 3\sin{x}$ \\ $ - 4\sin^3{x}$};
\node[circle, draw, below right = .5\mywidth and .5\mywidth of 2, inner sep=1pt, fill=blue!10](3){$\sin{x}$};
\node[circle, draw, below right = .2\mywidth and \mywidth of 3, inner sep=1pt, fill=teal!10](4){$\tan{x}=$ \\ $\sin{x}/\cos{x}$};
\node[circle, draw, above right = .5\mywidth and .5\mywidth of 4, inner sep=1pt, fill=green!10](5){$\tan{x}$};
\node[circle, draw, above left = .1\mywidth and .5\mywidth of 5, inner sep=1pt, fill=orange!10](6){$\cos{x}$};
\node[circle, draw, above left = .5\mywidth and .8\mywidth of 6, inner sep=1pt, fill=yellow!10](7){$\cos{x+y}=$ \\ $\cos{x}\cos{y}-$ \\ $\sin{x}\sin{y}$};
\node[circle, draw, above right = .8\mywidth and .2\mywidth of 6, inner sep=1pt, fill=lime!20](8){$\cos{\frac{\pi}{2}}$ \\ $=0$};

\draw[-latex] (1) edge[bend left] node[font=\tiny, inner sep=0pt, fill=white] {belongs to} (2);
\draw[-latex] (3) edge[bend right] node[font=\tiny, inner sep=0pt, fill=white] {$x=3x$} (2);
\draw[-latex] (3) edge[bend right] node[font=\tiny, inner sep=0pt, fill=white] {belongs to} (4);
\draw[-latex] (6) edge[bend right] node[font=\tiny, inner sep=0pt, fill=white] {belongs to} (4);
\draw[-latex] (5) edge[bend left] node[font=\tiny, inner sep=0pt, fill=white] {equals to} (4);
\draw[-latex] (6) edge[bend left] node[font=\tiny, inner sep=0pt, fill=white] {$x=x+y$} (7);
\draw[-latex] (6) edge[bend right] node[font=\tiny, inner sep=0pt, fill=white] {$x=\frac{\pi}{2}$} (8);

\end{tikzpicture}}
        \vspace{-0.5em}
		\caption{A subset of knowledge graph for trigonometry.} 
		\label{fig:kg}
 	\end{center}
  \vspace{-1.5em}
\end{figure}
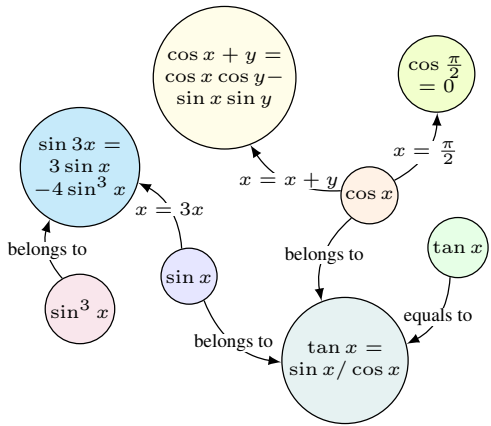

\subsection{Knowledge Graph Design and Exploration}
\label{ssec:kg}
\paragraph{Design.}
Figure~\ref{fig:kg} shows a subset of our KG. There are two types of nodes: (1) \textbf{Conceptual Node} and (2) \textbf{Theorem Node}. The Conceptual Node refers to foundational elements in trigonometry, for example, $\sin^3{x}$ and $\cos{x}$. 
On the other hand, Theorem Node represent general mathematical propositions or axioms, such as $\cos{\frac{\pi}{2}} = 0$. The edges between nodes represent the directional relationships or dependencies between the concepts. Structured as a directed graph, it features four types of connections: 
(1) \textbf{Dependency Link}, illustrating a concept's reliance on another. For instance, a link from the $\sin{x}$ node to $\tan{x} = \sin{x}/\cos{x}$ indicates the latter’s dependency on $\sin{x}$. (2) \textbf{Derivation Link}, signifying a derivation or logical progression from one concept or identity to another. An example is a link from $\sin{x}$ to $\sin{3x} = 3\sin{x} - 4\sin^3{x}$, indicating the derivation of the latter from the former. (3) \textbf{Application Link}, employed when a concept is utilized to deduce a specific instance or case. For instance, a link from $\cos{x}$ to $\cos(\frac{\pi}{2}) = 0$ demonstrates the application of $\cos{x}$ to a particular scenario. (4) \textbf{Identity Link}, which connects two identity nodes, demonstrating how one identity relates to or is transformed into another, such as $\tan{x}$ to $\tan{x} = \sin{x}/\cos{x}$.

\begin{algorithm}[t]
  \caption{Intermediate Result Selection}
  \label{alg:selection}
     \textbf{Input}: 1. Whole reasoning chains $\mathcal{C}$.  \\
     \hspace*{\algorithmicindent} \hspace*{1em} 2. List of deepest intermediate result state in \\ \hspace*{\algorithmicindent}\hspace*{1.4em} each reasoning chain: $\mathcal{L}_{all}$\\
    \textbf{Output}: Selected state list:  $\mathcal{L}_{sn}$ 
  \begin{algorithmic}[1]
  \State {N $\gets$ length($\mathcal{L}_{all}$)}
    \State $\mathcal{L}_{el}$ $\gets$ [ ] \textcolor{gray}{\Comment{// List to save eliminated state.}}
  \For{$i \gets 0$ to $N-1$}                    
        \For{$j \gets i+1$ to $N-1$}  
        \If{$\mathcal{L}_{all}$[j] in $\mathcal{L}_{el}$} Continue
        \EndIf
        \If{$\mathcal{L}_{all}$[i] in $\mathcal{L}_{el}$} Break
        \EndIf
        \State  \textcolor{gray}{\Comment // $s_{el}$ is the eliminated state or None.}
        \State $s_{el}$ $\gets$ PairWiseSelection($\mathcal{L}_{all}$[i], $\mathcal{L}_{all}$[j], $\mathcal{C}$)
        \If{$s_{el}$ != None}
        \State Add $s_{el}$ to $\mathcal{L}_{el}$
        \EndIf
    \EndFor
    \EndFor
    \State $\mathcal{L}_{sn}$ = $\mathcal{L}_{all}$ - $\mathcal{L}_{el}$ 
    \State \Return {$\mathcal{L}_{sn}$}
  \end{algorithmic}
\end{algorithm}
\vspace{-1em}
\paragraph{Information Retrieval.}
We employ in-context learning to distill features from the question, i.e., {function $E(\cdot)$}. Details of the query template can be found in Appendix~\ref{app:1}. The extracted features are categorized into two segments: (1) trigonometric functions e.g., $\sin{x}$, $\cos^2{x}$, $\sin{x}\cos{y}$, and (2) associated angles e.g., $\frac{\pi}{2}$, $3x$, $\pi - x$. Our approach to query the KG, {i.e., function $S(\cdot)$}, contains three steps. First, we temporarily exclude $\sin{x}$, $\cos{x}$, $\tan{x}$, $\sec{x}$, $\csc{x}$ and $\cot{x}$ from the extracted trigonometric functions. For the remaining functions, we search for nodes linked with them and retrieve their node information. Second, we integrate all extracted trigonometric functions with the involved angles. In this context, the trigonometric function serves to identify nodes, while the involved angles are used to match edges. Finally, we combine the data collected from both steps, removing any redundant details. This refined information is then utilized as contextual \textit{hints} within the prompt, aiding in solving the question.

As users apply our method to solve trigonometry problems, certain conclusions drawn from these problems can serve as lemmas for subsequent questions. Our KG is designed to be expandable. We offer an interface that allows users to add new nodes and edges to the KG. As more lemmas are incorporated, the KG becomes robust, providing more relevant information for future use.

\begin{figure*}[t]
	\begin{center}
		\subfloat[]{
   \includegraphics[width=0.2\textwidth]{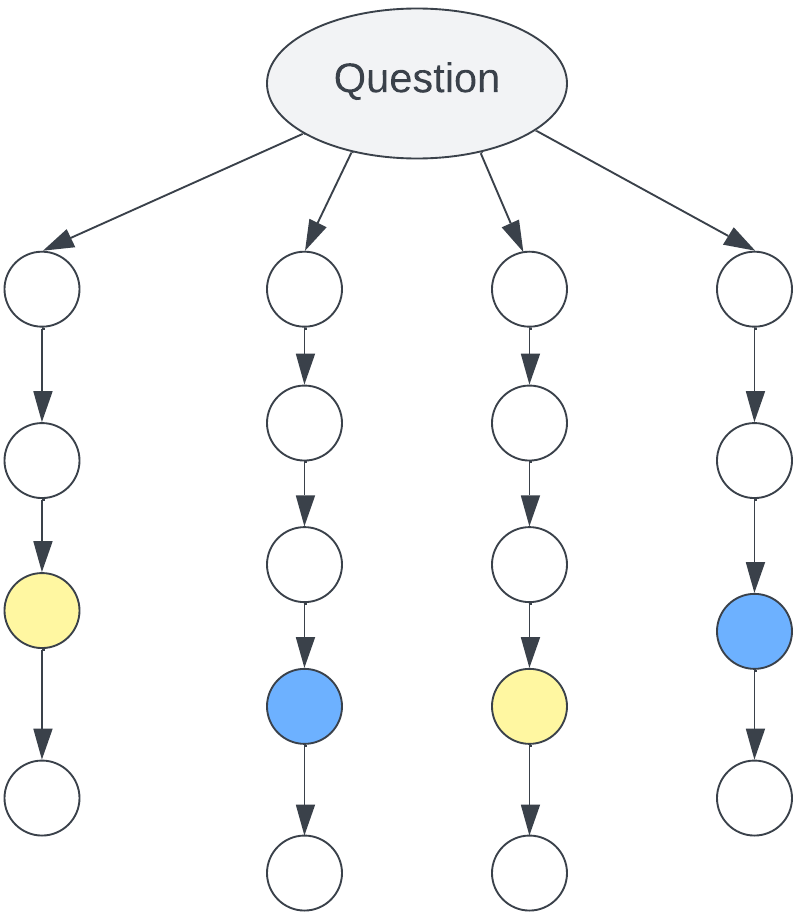}
			\label{fig:o1}
		}
  \hspace{0.2em}
		\subfloat[]{
  \includegraphics[width=0.2\textwidth]{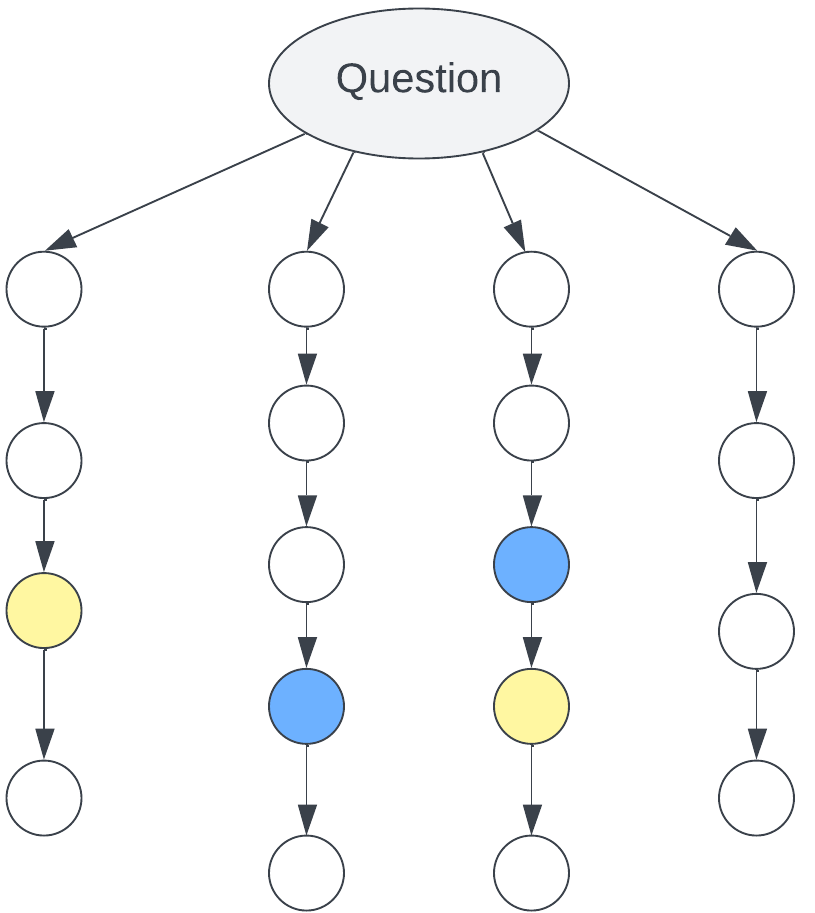}
			\label{fig:o2}
		}
  \hspace{0.2em}
		\subfloat[]{
  \includegraphics[width=0.2\textwidth]{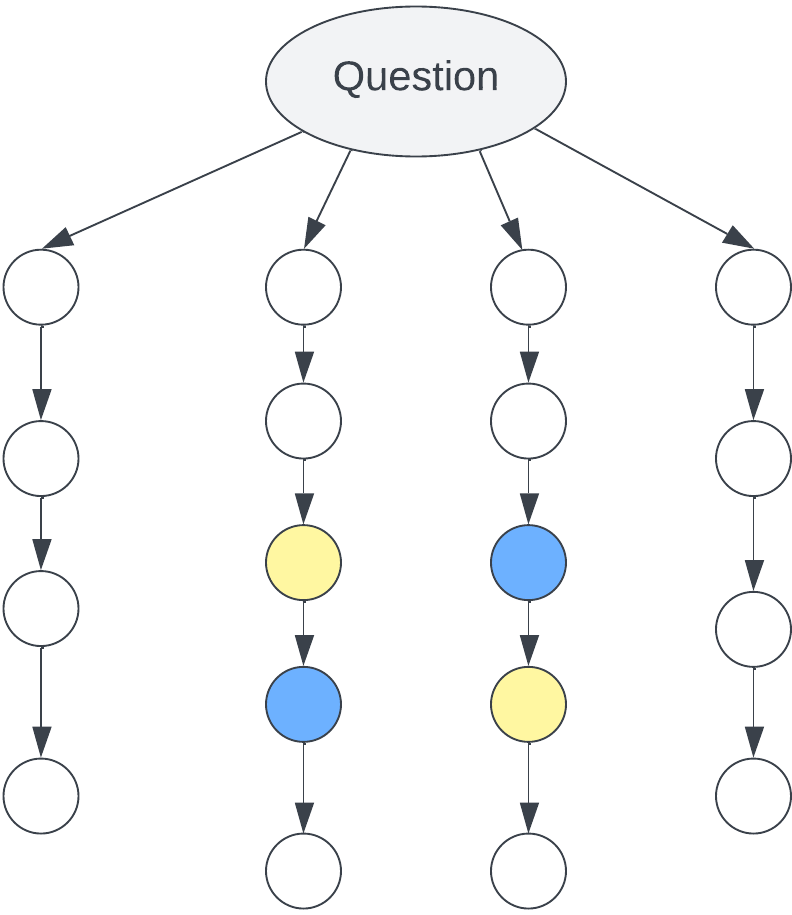}
			\label{fig:o3}
		}
  \hspace{0.2em}
		\subfloat[]{
  \includegraphics[width=0.2\textwidth]{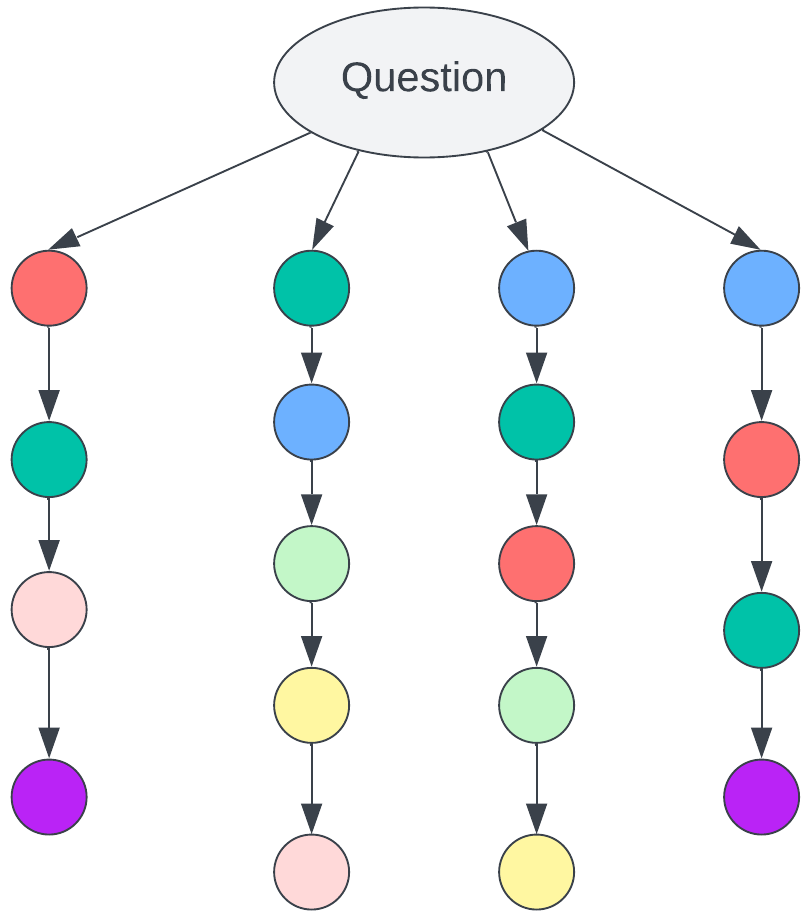}
			\label{fig:o4}
		}
  \vspace{-0.5em}
		\caption{Chains of thought with more than one group of overlapping intermediate result scenarios. (a) Two overlapping intermediate result groups, overlapping nodes in different chains. (b) Two overlapping intermediate result groups, nodes from different group appear in one chain. (c) Two overlapping intermediate result groups, nodes from different group appear in two chains. (d) More than two overlapping intermediate result groups.}
		\label{fig:overlapping_states}
 	\end{center}
  \vspace{-1.5em}
\end{figure*}

\subsection{Intermediate Result Selection}
\label{ssec:overlap}
In \alg, we posit that intermediate results in overlaps among multiple reasoning chains can be beneficial for subsequent reasoning stages. The intuition behind this is that errors in reasoning can manifest in various ways. But to accurately solve a problem, different methodologies are likely to converge on certain key intermediate results; 
Additionally, when we need to select a single state from multiple options within the same reasoning chain, the deepest state is chosen. This selection criterion is based on the observation that all states leading up to the deepest one contribute to its deduction, thereby making the selection of earlier states unnecessary for our purpose.
Based on these rationales, we explain how we arrive at $\mathbf{S}$, as illustrated in Figure~\ref{fig:workflow} for steps \circled{4} and \circled{8}. 

To identify overlapping intermediate results, \alg incorporates a step for assessing the similarity between pairs of results. This process involves converting each intermediate result into a TF (Term Frequency) - IDF (Inverse Document Frequency) vector. We then compute the pairwise cosine similarity for all vector pairs. If the cosine similarity between any two vectors exceeds the pre-set threshold $T$ ($T=0.999$), those intermediate results are deemed overlapping. TF-IDF and cosine similarity are widely utilized in information retrieval to calculate text similarity~\cite{schutze2008introduction}. 
Please note that our algorithm enables a \textit{human-in-the-loop} option (detail in Section~\ref{ssec:result}), as the overlapping states selection can be executed by human participants.
Upon all the overlapping intermediate results, we assess their respective priorities and choose those with the highest.
Given two groups defined as $A=\{x_{a_i}^{a_j}\}$ and $B=\{x_{b_i}^{b_j}\}$ and $A \cap B = \varnothing$. 
If $A$ can be inferred from $B$, then $B$ will not be selected as $A$ is more advanced in the reasoning chain. 
Therefore, we focus on the states from $A$ and $B$ that come from the same chain $m$, which we denote their positions in the chain as $a_j|_{a_i=m}$ and $b_j|_{b_i=m}$, respectively. 
{Chain number $m$ forms a set $M$.}
Concretely, the selected group after the comparison of $A$ and $B$ is as follows:
\begin{align}
\begin{cases}
B,  & \text{if $\forall m \in M$}, \text{$b_j|_{b_i=m} > a_j|_{a_i=m}$},
\\
A & \text{if $\forall m \in M$}, \text{$a_j|_{a_i=m} > b_j|_{b_i=m}$},
\\
    A \cup B, & \text{otherwise}.
  \end{cases}
  \end{align}

The first condition represents the scenario that the position of states in group $B$ is consistently deeper than those in $A$, indicating that $A$ is used to infer $B$. In this case, only the group $B$ will be kept. Similarly, $A$ is the inference of $B$ if the second condition fulfilled. The first two conditions are depicted in \Cref{fig:overlapping_states} (b). Beyond these two conditions, the two groups could be independent as depicted in \Cref{fig:overlapping_states} (a). In this scenario, $M=\varnothing$ indicating that there is no overlapping chain between two groups. \Cref{fig:overlapping_states} (c) illustrates that the two groups are intertwined. This can occur when simplifying a fractional expression: one might choose to simplify the numerator before the denominator, or vice versa. These steps can proceed in parallel but this simultaneity is not reflected in the reasoning chain.

\begin{table*}[h]
\centering
\resizebox{.9\linewidth}{!}{
\begin{tabular}{c|c|c|c|c|c|c|c|c}
\toprule[1pt]
Task & Algebra & \begin{tabular}{@{}c@{}} Counting \\  and Probability\end{tabular}   & Geometry & \begin{tabular}{@{}c@{}} Number \\  Theory\end{tabular}   &  Precalculus& Prealgebra & \begin{tabular}{@{}c@{}} Intermediate \\  Algebra\end{tabular}  &  \textbf{Total}\\ \hline
\#question & 307 & 123 &  132& 154 & 135 & 193 & 280 & 1324 \\ \midrule[1pt]
{\ttfamily{LLEMMA}}-7b &  9.1 &  4.1 & 2.3  & 3.9 &  2.2 & 9.8  & 2.1 &  5.3\\ \hline
{\ttfamily{LLEMMA}}-34b & 10.1  & 3.3 & 4.5 & 3.2 & 2.2 & 13.0 & 2.9 & 5.6 \\ \hline
ToT & 27.4  & 14.6 & 3.8& 11.0 & {1.5} & 34.7 & 3.2 & 15.3 \\ \hline
CoT-SC& 37.8 & 17.9 & 3.8 & 22.7  & 4.4 &  38.3& 3.2 & 20.2  \\ \midrule[1pt]
(Ours) SSC-CoT &   \textbf{42.7} & \textbf{25.2} & \textbf{15.9} & \textbf{31.8} & \textbf{7.4} & \textbf{49.7} & \textbf{8.9} & \textbf{27.4} \\ \bottomrule[1pt]
\end{tabular}
}
\caption{Result on MATH. Majority voting with $k$ = 20 is done for {\ttfamily{LLEMMA}} and CoT-SC. Best result is indicated in bold.}
\label{tab:result_table}
\end{table*}

Above mentioned scenarios only contain two overlapping groups, Figure~\ref{fig:o4} illustrates a more complex scenario which contains seven groups of overlapping intermediate results. The selection process for determining which group(s) of intermediate results to utilize is formalized in Algorithm~\ref{alg:selection}. 
In Algorithm~\ref{alg:selection}, we prepare (1) whole reasoning chains $\mathcal{C}$ and (2) 
list of deepest overlapping intermediate result state in each reasoning chain: $\mathcal{L}_{all}$. It is important to note that for each group of overlapping intermediate results, only one state will be included in $\mathcal{L}_{all}$. 
The \textit{PairWiseSelection} function, outlined in line 8, ascertains the relationship between any $\mathcal{L}_{all}[i]$ and $\mathcal{L}_{all}[j]$ within the reasoning chain, adhering to the rules previously defined for selection between two groups.
After completing all pairwise comparisons, the algorithm 
returns the final selected states. 

Finally, in scenarios where no overlap among intermediate results is observed, we will forward the final state from each reasoning chain to the verifier for further progression. 
If still no intermediate result are selected, the most recent set of intermediate results and related information will be reused for the current round of reasoning. 
However, each set of intermediate results and its related information is limited to a maximum of two uses.
Should there be no new overlapping intermediate results after two uses, we will revert to the set of intermediate results and related information that precedes the recently used one. This backtracking process will continue until we utilize the first set of overlapping intermediate results and related information.

\vspace{-10pt}
\section{Experiment}
\label{sec:exp}

In this section, we first introduce the experimental setup, including datasets, state-of-the-art baselines, and evaluation metrics. Then, we present the quantitative results, followed up with a qualitative result comparison to further demonstrate the advantage of our method.

\subsection{Experiment Setup}
\paragraph{Datasets.}
Our method is evaluated against other SOTA mathematical reasoning algorithms using the \data dataset. Additionally, we benchmark these methods using the MATH dataset~\cite{hendrycks2021measuring}. The MATH dataset comprises 12,500  questions from mathematics competitions, segmented into five ascending levels of difficulty, from level 1 to level 5. Given that our algorithm is specifically designed to tackle {complex} mathematical questions, our experiments focus exclusively on level 5 questions - the highest difficulty tier in MATH dataset. This level encompasses 1,324 questions spanning seven math categories: Algebra, Counting and Probability, Geometry, Number Theory, Precalculus, Prealgebra, and Intermediate Algebra. The number of questions in each category is given by \Cref{tab:result_table}.

\paragraph{Baselines.} 

In our research, we benchmark our \alg algorithm against other advanced in-context learning algorithms: Tree-of-Thought (ToT)~\cite{yao2023tree} and CoT-SC~\cite{wang2022self}. All three - ToT, CoT-SC, and our \alg - are implemented using the GPT-3.5.
Graph-of-Thought (GoT)~\cite{besta2023graph} is also a SOTA in-context multi-step reasoning algorithm.
However, GoT's design, which is not tailored specifically for mathematical reasoning, necessitates that input question be broken down into sub-tasks -- a requirement challenging to meet for mathematical questions. Therefore, GoT was not included as a baseline in our experiments.
Beyond in-context learning algorithms, our research also encompasses benchmarks against \lm~\cite{azerbayev2023llemma}, a language model explicitly developed for mathematical tasks and acclaimed for its SOTA performance across diverse mathematical datasets. Our experiments are conducted using both the 7B and 34B versions of the \lm model.

For \alg, we generate 5 reasoning chains per round, with a limit of 4 rounds, resulting in a maximum of 20 reasoning chains per question. In the case of ToT, we configure it to produce 5 steps at each level, selecting one for further development. We cap the LLM queries at 20 per question, counting only those queries that generate thoughts, excluding state evaluations using the LLM. For CoT-SC, we perform 20 queries per question to the foundational model and apply a majority vote mechanism on the outcomes. For the \lm experiments, we follow the CoT-SC procedure, with the only change being the replacement of the foundational model from the GPT-3.5 API to \lm.



Except for \alg, all other baselines use the same few-shot learning template, which is added in Appendix~\ref{app:base}. 
\alg cannot use the same template because \alg needs to extract intermediate results, therefore the expected output is different from other baselines.

\paragraph{Evaluation Metrics.} On the \data dataset, we compute scores for intermediate results as introduced in~\Cref{sec:dataset}. Specifically, we examine the deepest intermediate result correctly achieved by the model. The sum of scores over all questions will be used as the final result. Note that the maximum score on \data is 750. 
On MATH, we use the accuracy of answers generated by language models to indicate the model performance.

\subsection{Quantitative Results}
\label{ssec:result}
\begin{figure}[t]
	\begin{center}
			\includegraphics[width=0.7\columnwidth]{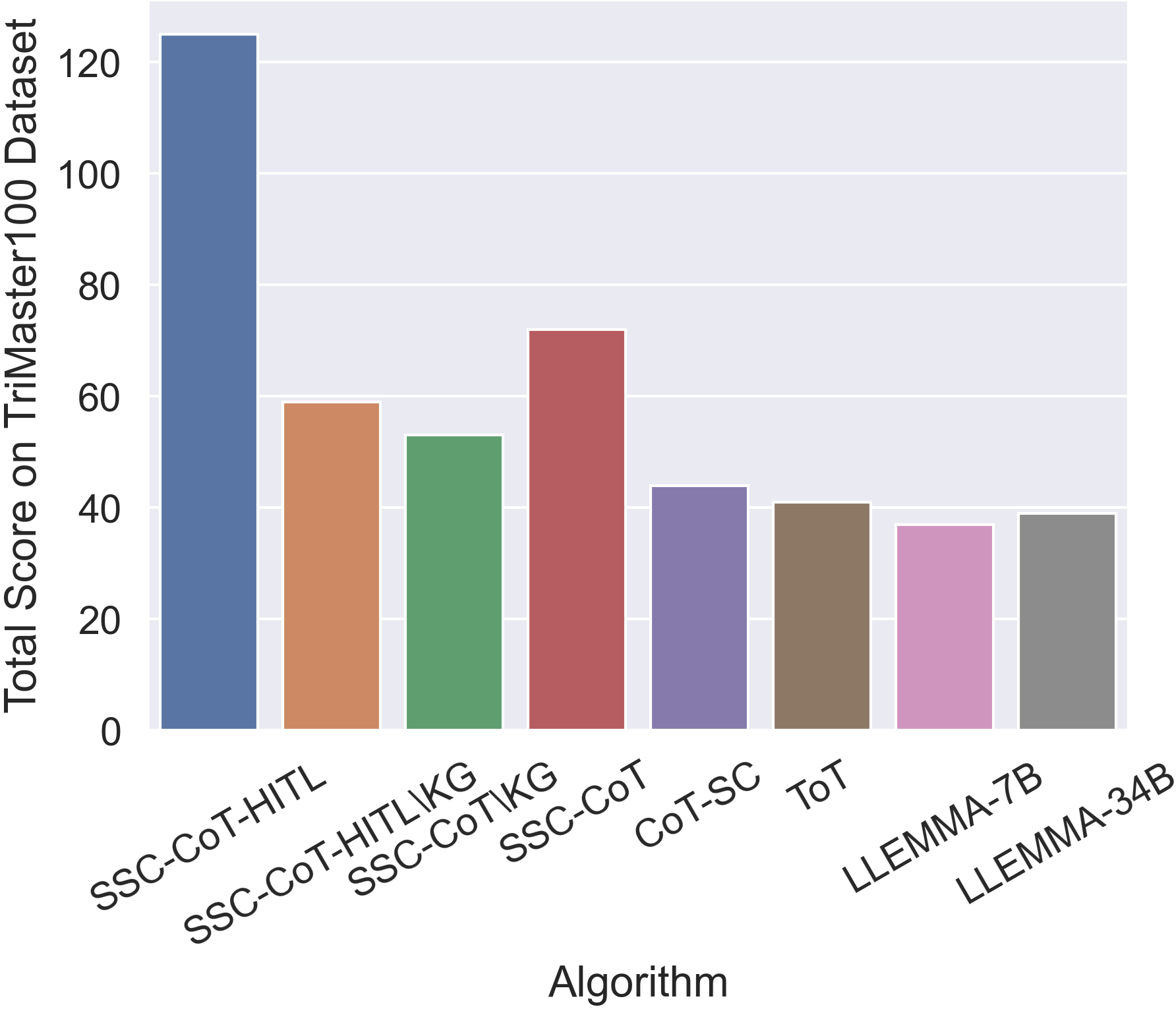}
   \vspace{-0.5em}
		\caption{Results on \data, full score is 750.} 
		\label{fig:tri_result}
 	\end{center}
  \vspace{-1.5em}
\end{figure}

\paragraph{Ablation Study.}
In this section, we evaluate the efficacy of key components in our \alg, KG and intermediate result selection, by introducing three variants: \textbf{SSC-CoT-HITL}, \textbf{SSC-CoT-HITL$\backslash$KG} and \textbf{SSC-CoT$\backslash$KG}. `HITL' denotes `human-in-the-loop', indicating that SSC-CoT-HITL incorporates human experts to select overlapping intermediate result, whereas SSC-CoT-HITL$\backslash$KG represents the SSC-CoT-HITL without KG.
{SSC-CoT$\backslash$KG} represents the variant that \alg without using KG.
Concretely, when comparing \alg and SSC-CoT-HITL with \alg$\backslash$KG and SSC-CoT-HITL$\backslash$KG in \Cref{fig:tri_result}, we see that utilizing KG efficiently improves the model's reasoning capabilities. Furthermore, the notable performance gap between SSC-CoT-HITL and \alg indicates a substantial boost from HITL intervention. When neither KG nor intermediate result selection is used, which is \sccot in \Cref{fig:tri_result}, SSC-CoT-HITL significantly outperforms, nearly tripling the score achieved by \sccot. Hence, KG and high quality intermediate result selection are both effective. 

\vspace{-0.5em}
\paragraph{Comparison with SOTA methods.}

In this section, we benchmark our \alg with baselines on both \data and MATH Level 5 to show its \textit{generalizability} in solving complex mathematical questions. On \textbf{\data}, our \alg in \Cref{fig:tri_result} surpasses all other baselines. For instance, \alg is $34\%$ higher than \sccot, the second best result. With a more general version of \alg, i.e. no KG is used, \alg{}$\backslash$KG is still $29\%$ higher than \sccot. The accuracy on final answer that all the baselines achieved on \data is discussed in Appendix~\ref{appendix:quantitative}.

On \textbf{MATH}, the comparison of our general method, \alg{}$\backslash$KG, with other baselines are shown in \Cref{tab:result_table}. Overall, \alg achieves the highest average accuracy and surpasses the second-best result by a large margin of $7.2\%$. 
Both \lm models show lower performance compared to other baseline models. This can be attributed to their relatively smaller size, resulting in reduced competence in understanding and reasoning compared to GPT-3.5.
It is worth noting that the performance across all algorithms is relatively better for Algebra and Prealgebra questions, reflecting the strengths and weaknesses of LLMs in processing different types of mathematical queries.

\subsection{Qualitative Results}
\Cref{tab:tri-result} demonstrates a solution provided by our \alg. It solves this question within $3$ rounds. With the related information in each round $r_k$ and $\mathbf{S}_v$ obtained from the last round, \alg is able to discover critical intermediates at each round, leading to the correct answer. ToT also focuses on simplifying $\sin{3A}$ to $\sin{A}$, but it makes a factual mistake that $\sin{3A}$ can be decomposed to $\sin{2A}\cdot \sin{A}$. This indicates that without related information, it is difficult for the model to deploy the correct identity. More qualitative results can be found in Appendix~\ref{appendix:qualitative}.
  \begin{table}
    \centering
     \resizebox{\linewidth}{!}{
     \begin{tabular}{{c}}
      \toprule
      $Q$:  Simplify $\sin{3A}/(1+2\cos{2A})$ \\
      \midrule \midrule 
         \textbf{Reasoning from \alg*} \\ 
         ------------------------------------------- \\
      \textbf{Round 1 - Given}: $Q$, $r_1$\\
     $\mathbf{S}_v$: \textcolor{blue}{$\sin{3A} = 3\sin{A} - 4\sin^3{A}$} \\
      \textbf{Round 2 - Given}: $Q$, $r_2$, $\mathbf{S}_v$\\
     $\mathbf{S}_v$: \textcolor{blue}{($3\sin{A} - 4\sin^3{A}$)} $/ (1+2\cos{2A})  =$ \textcolor{blue}{$\sin{A}(3-4\sin^2{A})$} $/ (1+2\cos{2A}) $ \\
\textbf{Round 3 - Given}: $Q$, $r_3$, $\mathbf{S}_v$\\
   $\mathbf{S}_v$:  \textcolor{olive}{$(1+2\cos{2A}) = 3 - 4\sin^2{A}$} \\
\textbf{Result}: $\sin{3A}/(1+2\cos{2A})= $\textcolor{blue}{($3\sin{A} - 4\sin^3{A}$)} $/$ \textcolor{olive}{($3 - 4\sin^2{A}$)} = $\sin{A}$ \\
      \midrule \midrule 
         \textbf{Reasoning from ToT} \\ 
         ------------------------------------------- \\
    \textbf{Step 1}: Use $\cos{2A} = 2\cos^2{A} - 1$ , $\sin{3A}/(1+2(2\cos^2{A}-1))$ \\
    \dots \\
    \textbf{Step 5}: Use identity $\sin^2{A} + \cos^2{A} = 1$ to rewrite the denominator as $4(1-\sin^2{A})-1$ \\
    \textbf{Step 6}: Factoring out a $\sin{A}$ in the numerator, we get \textcolor{red}{$\sin{A} \cdot \sin{2A}$} $/ (3-4\sin^2{A}$)\\
    \dots \\
    \bottomrule
    \end{tabular}
    }
    \raggedright
    \tiny{*In this problem, $r_1$ and $r_3$:\;`` \textbf{1.} $\sin{3A} = 3\sin{A} - 4\sin^3{A}$, \; \textbf{2.} $\cos{2A}$ = 1 - $2\sin^2{A}$, \; \textbf{3.} $\cos{2A} = 2\cos^2{A} - 1$". $r_2$:\;`` \textbf{1.} $\sin{3A} = 3\sin{A} - 4\sin^3{A}$." } 
    
    \caption{Solution provided by \alg (Top) and by ToT (Below) for a question from \data. Our \alg solves this question correctly based on $\mathbf{S}_v$ highlighted in colors, while ToT makes a factual mistake (in red) and cannot arrive at the correct answer.}\label{tab:tri-result}
    \vspace{-1em}
  \end{table}%

\section{Conclusion}

In this study, we introduce the Stepwise Self-Consistent Chain-of-Thought (\alg) algorithm, tailored for complex mathematical problem-solving. \alg improves the LLM's mathematical reasoning by identifying critical intermediate results through the intersection of diverse reasoning chains and integrating a knowledge graph.
To evaluate SSC-CoT, we collected a new dataset, \data, with 100 complex trigonometry questions, each divided into scored intermediate steps totaling 750 points. Results on \data and MATH level 5 datasets show \alg's superior performance compared to other methods, indicating its effectiveness in solving complex mathematical questions.
\vspace{-5pt}
\paragraph{Limitations and Future Work.} The use of human-in-the-loop intermediate result selection significantly enhances results in \Cref{fig:tri_result}. This indicates the potential for enhancing automatic detection of overlapping intermediate results. In our future work, we plan to deploy a reinforcement learning algorithm to train a reward model for selecting overlapping results based on human decisions. Moreover, the verification is conducted by querying the LLM with the given prompt. We believe that incorporating a more robust verifier, as discussed in~\cite{lets, dhuliawala2023chain}, has the potential to further improve \alg.
\vspace{-5pt}
\paragraph{Ethical Statement.} In this research, our goal is to improve the LLMs' capabilities to solve complex mathematical question. We also incorporated human experts' knowledge in some experiments to harvest an effective synergy between humans and AI models. 
To protect user privacy and rights, we have firmly followed the guidelines and regulations provided by our institution. 
We believe that by making AI more accessible, acceptable, and user-friendly, we can harness its potential to better assist humans.

\bibliography{main}
\bibliographystyle{icml2024}

\newpage
\appendix
\onecolumn
\section{Prompts.}

\subsection{Prompts on GPT-4 to construct the \data dataset}
\label{app:gpt}
For most of the case, to let GPT-4 to output reasoning steps, following prompt is enough.
\begin{tcolorbox}[colback=mybrown!5!white,colframe=mybrown!75!black]
Solve the question: \{\textbf{question}\}.
\end{tcolorbox}

For certain questions, GPT-4 may invoke an external component where a python code gets involved. In that case, the calculated result will directly output without reasoning steps. To avoid this, following prompt will be used for that type of question, for example the question: ``Find the value of $tan^2(20) + tan^2(40) + tan^2(80)$".

\begin{tcolorbox}[colback=mybrown!5!white,colframe=mybrown!75!black]
Solve the question: \{\textbf{question}\}. Please provide all steps of the reasoning. 
\end{tcolorbox}

\subsection{Prompt for $E$ of \alg to extract related information from question}
\label{app:1}

\begin{tcolorbox}[colback=mybrown!5!white,colframe=mybrown!75!black]
\textbf{Q:} For question: Simplify $\tan(100) + \sin(10)\cos(10) + (\cot(20))^2 + \sin(180+A)$. Extract trigonometric function and angles from the question. Be careful, for the pattern such as $\sin(10)\cos(10)$, we should extract the trigonometric function as $\sin(A)\cos(B)$, and for $(\cot(20))^2$, the extracted trigonometric function should be both $\cot(A)$ and $(\cot(A))^2$. There is no need to solve the problem, just provide the relevant information. \\
\textbf{A:} Trigonometric pattern(s): $\tan(A)$, $\sin(A)\cos(B)$, $\cot(A)$, $(\cot(A))^2$, $\sin(A)$. Angle(s): $100$, $10$, $180+A$.
\vspace{0.5em}

\textbf{Q:} For question: \{\textbf{question}\}. Extract trigonometric function and angles from the question. Be careful, for the pattern such as $\sin(10)\cos(10)$, we should extract the trigonometric function as $\sin(A)\cos(B)$, and for $(\cot(20))^2$, the extracted trigonometric function should be both $\cot(A)$ and $(\cot(A))^2$. There is no need to solve the problem, just provide the relevant information. \\
\textbf{A:}
\end{tcolorbox}

\subsection{Prompts for $V$ of \alg to verify of intermediate result}
\label{app:v}
Here, the function $V$ comprises two steps. Initially, we employ the following prompt to generate the entire reasoning process for determining the correctness of intermediate results for solving the given question.
\begin{tcolorbox}[colback=mybrown!5!white,colframe=mybrown!75!black]
Given the mathematical question: \{\textbf{question}\}, we have following inferences: \{\textbf{intermediate results}\}. Do you think it is correct? Let's think step by step.
\end{tcolorbox}
After getting the entire process, we give the whole reasoning process to the following prompt to conclude whether the the intermediate results are correct.

\begin{tcolorbox}[colback=mybrown!5!white,colframe=mybrown!75!black]
We have the inference as follows: \{\textbf{inference process}\}. According to the inference, please conclude a clear answer. Please only output ``yes" or ``no".
\end{tcolorbox}




\subsection{Prompt for $\mathcal{G}_1$ of \alg to generate reasoning chain}
\label{app:4}
\begin{tcolorbox}[colback=mybrown!5!white,colframe=mybrown!75!black]
    \textbf{Q.} Simplify
    \[
    \frac{-\sin(180+A) + \sin(-A) - \tan(360 + A)}{\tan(A+180) + \cos(-A) + \cos(180-A)}.
    \]
    Useful knowledge:
    \begin{enumerate}
        \item $\sin(-X) = -\sin X$.
        \item $\sin(180+X) = -\sin X$.
        \item $\tan(180 + X) = \tan X$.
        \item $\cos(180-X) = -\cos X$.
        \item $\cos(-X) = \cos X$.
    \end{enumerate}
    In the end of reasoning, we have to summarize reasoning into concrete intermediate results without explanation (pure equation without extra text). 
    
    \textbf{A.} Let's simplify the expression
    \[
    \frac{-\sin(180+A) + \sin(-A) - \tan(360 + A)}{\tan(A+180) + \cos(-A)+ \cos(180-A)}.
    \]
    For the numerator part:
    \[
    -\sin(180+A) + \sin(-A) - \tan(360 + A) = -\tan(A).
    \]
    For the denominator part:
    \[
    \tan(A+180) + \cos(-A)+ \cos(180-A) = \tan(A).
    \]
    Therefore,
    \[
    \frac{-\sin(180+A) + \sin(-A) - \tan(360 + A)}{\tan(A+180) + \cos(-A)+ \cos(180-A)} = \frac{-\tan(A)}{\tan(A)} = -1.
    \]
    To summarize, we can conclude our reasoning into three intermediate results:
    \begin{enumerate}
        \item $-\sin(180+A) + \sin(-A) - \tan(360 + A) = -\tan(A)$,
        \item $\tan(A+180) + \cos(-A)+ \cos(180-A) = \tan(A)$,
        \item $\frac{-\sin(180+A) + \sin(-A) - \tan(360 + A)}{\tan(A+180) + \cos(-A)+ \cos(180-A)} = \frac{-\tan(A)}{\tan(A)} = -1$.
    \end{enumerate}
    
    \textbf{Q.} \{\textbf{question}\} \\
    Useful knowledge: \{\textbf{related information}\} \\
    In the end of reasoning, we have to summarize reasoning into concrete intermediate results without explanation (pure equation without extra text). 
    
    \textbf{A.}
\end{tcolorbox}

\subsection{Prompt for $\mathcal{G}$ of \alg to generate reasoning chain}
\label{app:ssc_g}

\begin{tcolorbox}[colback=mybrown!5!white,colframe=mybrown!75!black]
    \textbf{Q.} Simplify
    \[
    \frac{-\sin(180+A) + \sin(-A) - \tan(360 + A)}{\tan(A+180) + \cos(-A) + \cos(180-A)}.
    \]
    Intermediate result 1:
    \[
    \begin{aligned}
        & \frac{-\sin(180+A) + \sin(-A) - \tan(360 + A)}{\tan(A+180) + \cos(-A) + \cos(180-A)} \\
        & = \frac{\sin(A) - \sin(A) - \tan(360 + A)}{\tan(A+180) + \cos(A) - \cos(A)}.
    \end{aligned}
    \]
    Useful knowledge:
    \begin{enumerate}
        \item $\sin(-X) = -\sin X$.
        \item $\sin(180+X) = -\sin X$.
        \item $\tan(180 + X) = \tan X$.
        \item $\cos(180-X) = -\cos X$.
        \item $\cos(-X) = \cos X$.
    \end{enumerate}
    \textbf{A.} According to the intermediate result, simplify the expression
    \[
    \frac{-\sin(180+A) + \sin(-A) - \tan(360 + A)}{\tan(A+180) + \cos(-A) + \cos(180-A)}
    \]
    to
    \[
    \frac{\sin(A) - \sin(A) - \tan(360 + A)}{\tan(A+180) + \cos(A) - \cos(A)}.
    \]
    For the numerator part:
    \[
    \sin(A) - \sin(A) - \tan(360 + A) = -\tan(A).
    \]
    For the denominator part:
    \[
    \tan(A+180) + \cos(A) - \cos(A) = \tan(A).
    \]
    Therefore,
    \[
    \frac{\sin(A) - \sin(A) - \tan(360 + A)}{\tan(A+180) + \cos(A) - \cos(A)} = \frac{-\tan(A)}{\tan(A)} = -1.
    \]
    To summarize, we can conclude our reasoning into three intermediate results:
    \begin{enumerate}
        \item $\sin(A) - \sin(A) - \tan(360 + A) = -\tan(A)$,
        \item $\tan(A+180) + \cos(A) - \cos(A) = \tan(A)$,
        \item $\frac{\sin(A) - \sin(A) - \tan(360 + A)}{\tan(A+180) + \cos(A) - \cos(A)} = \frac{\sin(A)}{\tan(A)} = -1$.
    \end{enumerate}
    \textbf{Q.} \{\textbf{question}\}. \\
    Intermediate result: \{\textbf{intermediate results}\}. \\
    Useful knowledge: \{\textbf{related information}\}. \\
    In the end of reasoning, we have to summarize reasoning into concrete intermediate results without explanation (pure equation without extra text).
    
    \textbf{A.}
\end{tcolorbox}

\subsection{Prompt of few-shot learning for ToT, CoT-SC and \lm}
\label{app:base}

\begin{tcolorbox}[colback=mybrown!5!white,colframe=mybrown!75!black]
\textbf{Example:}
\begin{enumerate}
  \item \textbf{Question:} Consider the function \( f(x) \) defined as follows: \( f(x) = ax+3 \) if \( x > 2 \), \( x-5 \) if \( -2 \leq x \leq 2 \), and \( 2x-b \) if \( x < -2 \). Find \( a+b \) if the function is continuous.\\
  \textbf{Solution:}
  \begin{enumerate}
    \item \textbf{Step 1:} Ensure continuity at \( x=2 \) and \( x=-2 \). Equate \( ax+3 \) and \( x-5 \) at \( x=2 \).
    \item \textbf{Step 2:} Solving \( a(2)+3 = 2-5 \) leads to \( 2a = -6 \), so \( a = -3 \).
    \item \textbf{Step 3:} Equate \( x-5 \) and \( 2x-b \) at \( x=-2 \).
    \item \textbf{Step 4:} Solving \( -2-5 = 2(-2)-b \) gives \( b = 3 \). Therefore, \( a+b = -3+3 \).
  \end{enumerate}
  \textbf{Final answer:} 0.
\end{enumerate}

\textbf{Example:}
\begin{enumerate}
  \item \textbf{Question:} What number is \( 64\% \) of 16?\\
  \textbf{Solution:}
  \begin{enumerate}
    \item \textbf{Step 1:} Let the number be \( x \). Set up the equation \( \frac{16}{x} = \frac{64}{100} \).
    \item \textbf{Step 2:} Simplify to \( \frac{1}{x} = \frac{4}{100} = \frac{1}{25} \), so \( x = 25 \).
  \end{enumerate}
  \textbf{Final answer:} 25.
\end{enumerate}

\textbf{Example:}
\begin{enumerate}
  \item \textbf{Question:} Given three complex numbers \( a+bi \), \( c+di \), \( e+fi \), with \( b=1 \), \( e=-a-c \), and their sum equal to \( -i \), find \( d+f \).\\
  \textbf{Solution:}
  \begin{enumerate}
    \item \textbf{Step 1:} Sum the complex numbers: \( a+bi + c+di + e+fi = -i \). The real parts sum to 0, and imaginary parts sum to -1.
    \item \textbf{Step 2:} The equations become \( a+c+e=0 \) and \( b+d+f=-1 \).
    \item \textbf{Step 3:} With \( b=1 \), solve for \( d+f \), getting -2.
  \end{enumerate}
  \textbf{Final answer:} -2.
\end{enumerate}

\textbf{Task:}
\begin{enumerate}
  \item \textbf{Question:} \{\textbf{question}\} Think step by step and explain the reasoning for the final answer like the examples. Only include the current step number and explanation in your answer. Do not repeat the question or previous steps.\\
  \textbf{Solution:}
\end{enumerate}

\end{tcolorbox}

\section{Additional Quantitative Result}
\label{appendix:quantitative}
In Table~\ref{table:accuracy}, we present the accuracy metrics for all baseline models applied to the \data datasets. Notably, the \lm 7B model completely solves 2 questions. Judging purely by accuracy, \lm 7B surpasses CoT-SC, ToT, SSC-CoT$\backslash$KG, and \lm 34B. However, a closer examination of the results in Table~\ref{tab:tri-result} reveals that \lm 7B ranks lowest overall. This discrepancy in performance metrics underscores our initial assertion that for intricate mathematical questions, the rarity of complete solutions diminishes the significance of accuracy as a measure of an algorithm's mathematical reasoning capabilities. Consequently, our \data dataset, which includes scored intermediate results, offers a more appropriate benchmark for evaluating the proficiency in complex mathematical question reasoning.
\begin{table*}[ht]
\centering

\begin{tabular}{c c c c c c c c c}
\hline
\textbf{SSC-CoT-HITL} &  \textbf{SSC-CoT-HITL$\backslash$KG} & \textbf{SSC-CoT$\backslash$KG}& \textbf{\alg}&  \textbf{CoT-SC} &  \textbf{ToT}  & \textbf{\lm 7B}  & \textbf{\lm 34B}\\
\hline
\textbf{5} & 2&1&2&1&1&2&1\\
\hline
\end{tabular}
\caption{Accuracy (\%) Result of All Baselines on \data.}
\label{table:accuracy}
\end{table*}

\section{Additional Qualitative Result}
\label{appendix:qualitative}
Table~\ref{table:app_quali} presents the mathematical reasoning process for an algebra question from MATH level 5. Initially, both algorithms correctly infer the first step. However, from the second round onward, our method, which utilizes an overlapping-based selection mechanism for intermediate results, successfully identifies and retains the crucial intermediate results. The verification process ensures the accuracy of the selected states, notably $16-4c < 0$, guiding us to the correct conclusion. In contrast, ToT only arrives at the expression $16-4c$ in its third step, displaying a slower pace compared to \alg. Furthermore, an incorrect inference at the second step by ToT leads to a cascade of erroneous conclusions. This outcome demonstrates that \alg not only identifies critical intermediate results more quickly but also selects the accurate ones through verification steps.
\begin{table}[H]
    \centering
    
    \resizebox{\linewidth}{!} {
    
    \begin{tabular}{{c}}
    \toprule
    $Q$:  What is the smallest integer value of $c$ such that the function $f(x)=\frac{2x^2+x+5}{x^2+4x+c}$ has a domain of all real numbers? \\
    \midrule \midrule 
        \textbf{Reasoning from SSC-CoT} \\ 
        ------------------------------------------- \\
        \textbf{Round 1 - Given}: $Q$ \\
        $\mathbf{S}_v$: {$x^2+4x+c$} != {$0$}  \\

        \textbf{Round 2 - Given}: $Q$, $\mathbf{S}_v$\\
        $\mathbf{S}_v$: {$16-4c < 0$}   \\

        \textbf{Round 3 - Given}: $Q$, $\mathbf{S}_v$\\
        $\mathbf{S}_v$:  {$c > 4$}  \\

        \textbf{Result}: $c = 5$ \\
    \midrule \midrule 
    
    \textbf{Reasoning from ToT} \\ 
------------------------------------------- \\

    \textbf{Step 1}: We get a conclusion that $x^2+4x+c$ should not equal 0.\\
    \textbf{Step 2}: We further get a conclusion that \textcolor{red}{$b^2-4ac>0$}\\
    \textbf{Step 3}: After putting the values of a and b into $b^2-4ac$ respectively, we get \textcolor{red}{$16-4c > 0 $}\\
    \textbf{Step 4}: Finally, we get  a conclusion :\textcolor{red}{$c < 4$}\\
    \textbf{Step 5}: So, the answer is \textcolor{red}{$c = 4$}\\
    \bottomrule
    \end{tabular}
    }
    \caption{Solution from \alg (Top) and from ToT (Below) for an algebra question from MATH level 5. Mistakes during reasoning are highlighted in red.}
    \label{table:app_quali}
\end{table}%

Table~\ref{table:app_quali2} illustrates the mathematical reasoning process for another algebra question at MATH level 5. In this instance, \alg makes an error, yet ToT eventually resolves the question, highlighting the challenges of applying \alg to LLMs. During the first round, \alg partially solves the question. However, the second round reveals disorganized calculations for $k$, producing several values without any overlap. $x$ value is also recalculated, which is incorrect and leads to an erroneous conclusion. This underscores the known difficulty of GPT-3 in performing arithmetic tasks~\cite{app14020744}. Specifically, in round 2, two types of errors are observed: incorrect calculation of $k$ and $x$ despite accurate values for $a$, $b$, $c$, and $x$ are given, and a verification process that fails to accurately assess intermediate results, mistakenly using $x = -1$ as $S_v$. As we mentioned the limitations in the Conclusion suggests that enhancing verification could improve \alg's performance. Nevertheless, it is worth noting that in round 1, \alg accurately determined the values of $a$, $b$, $c$, and $x$. The same results are only found at step 3 by ToT, thereby reaffirming \alg's capability in quickly pinpointing critical intermediate results.

\begin{table}[H]
    \centering
    
    \resizebox{\linewidth}{!} {
    
    \begin{tabular}{{c}}
    \toprule

    $Q$:  In an equation of the form $k = ax^2 + bx + c$ with $a > 0$, the least possible value of $k$ occurs at $x = \frac{-b}{2a}$. \\ In the equation $k = (6x + 12)(x - 8)$, what is the least possible value for $k$? \\
    \midrule \midrule 
        \textbf{Reasoning from SSC-CoT} \\ 
        ------------------------------------------- \\
	
        \textbf{Round 1 - Given}: $Q$ \\
        $\mathbf{S}_v$: {$a=6$}, {$b=-36$}, {$c=-96$}, {$x=3$} \\    

        \textbf{Round 2 - Given}: $Q$, $\mathbf{S}_v$\\
        $\mathbf{S}_v$: \textcolor{red}{$x=-1$}  \\

        \textbf{Round 3 - Given}: $Q$, $\mathbf{S}_v$\\
        $\mathbf{S}_v$:  \textcolor{red}{$k=54$}   \\

        \textbf{Result}: $ \textcolor{red}{k = 54}$ \\
    \midrule \midrule 
    
    \textbf{Reasoning from ToT} \\ 
------------------------------------------- \\
    \textbf{Step 1}: From the question, we can determine that $k = (6x+12)(x-8)$\\
    \textbf{Step 2}: From $k = (6x+12)(x-8)$, we know that $k = 6x^{2}-48x+12x-96$ \\
    \textbf{Step 3}: We can get that $k = 6x^{2}-36x-96$ so that $a=6$, $b=-36$ and $c=-96$\\
    \textbf{Step 4}: Due to $x=\frac{-b}{2a}$, we can get a conclution that $x=3$\\
    \textbf{Step 5}:By bringing the result of $a$, $b$, $c$ and $x$ into $k=ax^{2}+bx+c$, we can conclude that $k=-150$\\
    \bottomrule
    \end{tabular}
    }
    \caption{Solution from \alg (Top) and from ToT (Below) for an algebra question from MATH level 5. Mistakes during reasoning are highlighted in red.}
    \label{table:app_quali2}
\end{table}%

\section{Computational Infrastructure Details}
\label{supp:infrastructure}
All experiments in this paper are conducted on the device given in \Cref{supp-tab:device}.
\begin{table}[h]
\centering

\label{supp-tab:device}
\begin{tabular}{c|c}
\toprule[1pt]
Device Attribute  & Value \\\hline 
Computing infrastructure&     GPU              \\
GPU model &    NVIDIA A100              \\
GPU number  &     1                   \\
CUDA version   &     12.2             \\
\bottomrule[1pt]
\end{tabular}
\caption{Computational infrastructure details.}
\end{table}


\end{document}